\documentclass[letterpaper, 10 pt, conference]{ieeeconf}  % Comment this line out if you need a4paper

\IEEEoverridecommandlockouts                              % This command is only needed if 
\usepackage{cite}
\usepackage{graphics} % for pdf, bitmapped graphics files
\usepackage{epsfig} % for postscript graphics files
\usepackage{mathptmx} % assumes new font selection scheme installed
\usepackage{graphicx}
\usepackage[english]{babel}
\usepackage{blindtext}
\usepackage{times} % assumes new font selection scheme installed
\usepackage{amsmath} % assumes amsmath package installed
\usepackage[ruled,norelsize]{algorithm2e}
\usepackage{algpseudocode}
\usepackage{url}
\usepackage{amssymb}  % assumes amsmath package installed
\usepackage{epstopdf}
\usepackage{todonotes}
\usepackage{lipsum}
\usepackage{subcaption}
\usepackage{bbm}

\title{\LARGE \bf Bayesian Optimization Meets Laplace Approximation \\ for Robotic Introspection}
\author{Matthias~Humt$^{1}$, Jongseok~Lee$^{1}$ and Rudolph~Triebel$^{1, 2}$% <-this % stops a space
\thanks{$^{1}$ Institute of Robotics and Mechatronics, German Aerospace Center (DLR), Wessling, Germany. {\tt\small matthias.humt@dlr.de}}%
\thanks{$^{2}$ Computer Vision Group, Technical University of Munich, Garching, Germany}
}

\begin{document}

\maketitle
\thispagestyle{empty}
\pagestyle{empty}

%%%%%%%%%%%%%%%%%%%%%%%%%%%%%%%%%%%%%%%%%%%%%%%%%%%%%%%%%%%%%%%%%%%%%%%%%%%%%%%%
\begin{abstract}
In robotics, deep learning (DL) methods are used more and more widely, but their general inability to provide reliable confidence estimates will ultimately lead to fragile and unreliable systems. This impedes the potential deployments of DL methods for long-term autonomy. Therefore, in this paper we introduce a scalable Laplace Approximation (LA) technique to make Deep Neural Networks (DNNs) more introspective, i.e. to enable them to provide accurate assessments of their failure probability for unseen test data. In particular, we propose a novel Bayesian Optimization (BO) algorithm to mitigate their
tendency of under-fitting the true weight posterior, so that both the calibration and the accuracy of the predictions can be simultaneously optimized. We demonstrate empirically that the proposed BO approach requires fewer iterations for this when compared to random search, and we show that the proposed framework can be scaled up to large datasets and architectures.
\end{abstract}

%%%%%%%%%%%%%%%%%%%%%%%%%%%%%%%%%%%%%%%%%%%%%%%%%%%%%%%%%%%%%%%%%%%%%%%%%%%%%%%%

\section{Introduction}
Space exploration, maintenance $\&$ inspection and surveillance $\&$ reconnaissance are just a few examples of application areas where robots are needed that are capable of long-term autonomous operation. One of the crucial aspects of achieving such long-term autonomy is the addition of the cognitive ability to doubt and understand their own failures and further assess the quality of their own internal states such as sensor data, predictions of a learning method, representation of the world etc. Such knowledge, when examined accurately, can alter the behavior of robots by including the assessed reliability score into the planning and mission management process so that higher mission success rates can be achieved. Such cognitive abilities are called robotic introspection \cite{grimmett2013knowing}.

For Deep Neural Networks (DNNs), the quality of the learned predictor can be assessed by generating distributions rather than a single most-likely prediction. There exists ample evidence that such confidence estimates, if well-calibrated w.r.t the accuracy of the learning algorithm, can lead to improved decision-making and failure avoidance \cite{grimmett2016introspective, triebel2016driven, mcallister2017concrete}. Unfortunately, obtaining these confidence estimates for DNNs is often subject to a fidelity-scalability trade-off; highly scalable methods such as MC-dropout \cite{gal2016dropout} often underestimate predictive uncertainty \cite{li2017dropout} while sampling methods such as HMC \cite{neal2012bayesian} are widely recognized as gold standard but do not scale to large datasets and modern DNN architectures. As the requirements vary depending on the use-case, we hypothesize that an easy-to-use method that can provide a first line-of-attack with a balanced fidelity-scalability would be beneficial for save long-term autonomous operation in robotics.
\begin{figure}[t]
    \centering
    \includegraphics[width=\linewidth]{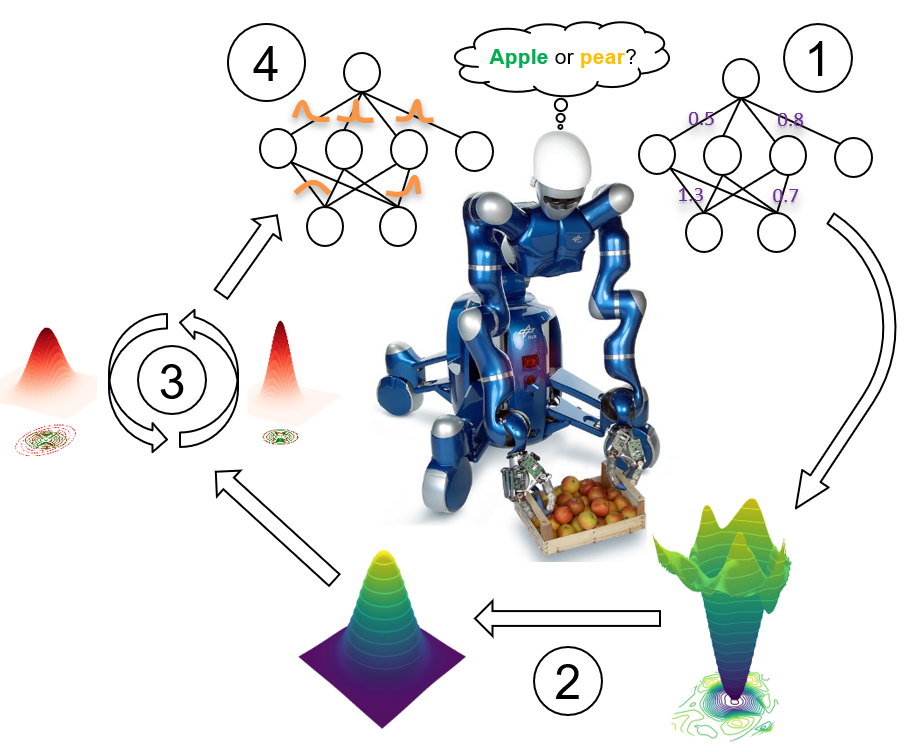}
    \caption{To equip our robots with introspective qualities, we transform a Deep Neural Network (DNN) (1) into a Bayesian Neural Network (4) using Laplace Approximation (2) which employs information from the inverse curvature (Hessian) of the loss landscape of the DNN (2, right) for the covariance matrix of a multivariate Gaussian as an approximation to the DNNs weight posterior distribution (2, left) and mitigate its tendency of underfitting through Bayesian Optimization (3).}
    \label{fig:la}
\end{figure}
For this purpose, we introduce a scalable LA technique \cite{mackay1992practical, ritter2018scalable} in the context of robotic introspection. As depicted in Fig. \ref{fig:la}, LA computes the DNNs posterior with a Gaussian distribution around a local minimum, where its mean is the maximum a-posteriori (MAP) estimate of the DNNs parameters (i.e. the trained weights) while the covariance matrix is specified by the scaled inverse of the DNNs Hessian matrix (i.e. the matrix of 2nd order derivatives of the loss w.r.t the weights). Here, the Hessian matrix can be obtained via 2nd order Taylor series expansion on the log-likelihood and requires relaxed assumptions on the DNN architecture (e.g. no mandatory dropout layers). By exploiting Kronecker factorization of the Hessian, more expressive posterior families than the Bernoulli distribution or diagonal approximations of the covariance matrix can further be modelled, even in a large scale setting that is relevant for many robotic applications \cite{lee2020estimating}. This was demonstrated in \cite{shinde2020learning}, where LA induced uncertainty was successfully employed to improve robustness and accuracy of deep sensor fusion for end-to-end trainable visual-intertial odometry.

More importantly, we propose a new Bayesian Optimization (BO) technique to address the under-fitting problem of LA which is its inherent drawback arising from (i) the use of 2nd order Taylor series expansion on the DNNs local log-likelihood restricting the assumed posterior to be a bell-shaped distribution (i.e. it cannot handle an asymmetrical true posterior and places a large probability mass in low probable sets of parameters), and (ii) a typical over-parameterization of DNNs which result in an undetermined Hessian matrix. Opposed to the brute force grid search strategy used in \cite{ritter2018scalable}, we empirically show that the proposed BO technique can provide faster convergence for better calibration and accuracy of the predictions. In addition, as we show that the proposed framework can be scalably applied to ImageNet, we demonstrate an alternative approach to MC-dropout, which is a standard approach to obtaining model (i.e. epistemic) uncertainty in DNNs.

In summary, our contributions are:
\begin{itemize}
    \item We propose to employ a scalable yet straight forward Laplace Approximation technique in the context of robotic introspection.  
    \item We propose a Bayesian Optimization technique to tackle the under-fitting problem of Laplace Approximation.
    \item We demonstrate empirically that the proposed technique requires fewer iterations when compared to a random search, and the resulting framework scales to large network architectures and datasets.
\end{itemize}
\section{Related Work}
Uncertainty estimation in DNNs has received increasing attention in various communities and a variety of methods have been proposed. Broadly, we build upon the paradigm of Bayesian Deep Learning (BDL) which aims to estimate the posterior distribution over the weights of DNNs, which allows for predictive uncertainty to be computed by marginalization over the possible values of weights \cite{mackay1992practical}. MC-dropout \cite{gal2016dropout} is a prominent example within this umbrella, where the authors showed that training a DNN with dropout layers is in some sense equivalent to performing variational inference - an optimization based approach to estimating the posterior distribution rather than sampling. While MC-dropout has been widely adopted in the robotics community and beyond \cite{feng2019introspective, miller2018dropout, kahn2017uncertainty}, some drawbacks of the framework include (i) the need for dropout layers while most modern networks are trained with batch normalization, and (ii) lack of expressiveness of the posterior family. Concurrently, more sophisticated methods based on sampling \cite{neal2012bayesian} and variational inference \cite{Hinton1993} have been widely studied for DNNs. However, these arguably complex Bayesian inference procedures have not been popular as they are often not scalable or fragile \cite{wu2018deterministic}. We instead build upon LA \cite{mackay1992practical, ritter2018scalable}, one of the simplest and deterministic inference methods with an attempt to make the framework more practical for real-world robotics applications. We are aware that there exists alternatives to BDL, namely post-hoc calibration methods \cite{wenger2019non, guo2017calibration}, ensembles \cite{lakshminarayanan2017simple}, probabilistic programming \cite{tran2016edward, bingham2019pyro} etc. but these approaches are rather orthogonal to our work. 

The importance of robotic introspection for decision making has been first emphasized in \cite{grimmett2013knowing}, where the authors demonstrated that on top of standard metrics for classification, namely precision and recall, the quality of confidence estimates are an additional crucial measure that can help to avoid catastrophic failures of robots, and can also be utilized for active learning \cite{triebel2016driven}. Since then, \cite{daftry2016introspective, ramanagopal2018failing, guruau2018learn} have proposed to learn the failures of a perception module with a DNN, and \cite{saxena2017learning} demonstrated a failure recovery method. The backbone of these methods is to perform extensive data collection in the regime where the perception module fails, and use such data for training a DNN based failure predictor. Coping with non-stationary data \cite{grimmett2013knowing} (also called open-set conditions in \cite{miller2018dropout}) by means of unsupervised novelty detection is a further notable class of methods that attempted to achieve robotic introspection \cite{richter2017safe, wong2019identifying}. Our method relies on Bayesian principles and is not tied to a specific task but can also be applied to pruning, active learning, continual learning, and 2nd order optimization for DNNs, which are other crucial frontiers towards long-term autonomy.
\section{Bayesian Hyperparameter Optimization for Laplace Approximation }

\subsection{Laplace Approximation}
Let $f_{\mathbf{W}}(\mathbf{x})$ be a DNN with weights $\mathbf{W}$ that transforms inputs $\mathbf{x}$ into outputs $\hat{\mathbf{y}}$. Here, $\mathbf{x}$ are images and $\hat{\mathbf{y}}$ are predicted class probabilities with corresponding ground truth labels $\mathbf{y}$.
After training using the data $\mathcal{D}=\{\mathbf{x}_i,\mathbf{y}_i\}_{i=1}^N$, the weights $\mathbf{W}^\star$ are the mode of the networks posterior probability distribution which can be approximated by a 2nd order Taylor expansion around this mode which is commonly referred to as Laplace Approximation (LA),
\begin{equation}\label{eq:taylor}
    \log p(\mathbf{W}|\mathcal{D})\approx \log p(\mathbf{W}^\star|\mathcal{D})-\frac{1}{2}(\mathbf{W}-\mathbf{W}^\star)^T H(\mathbf{W}-\mathbf{W}^\star),
\end{equation}
where $H$ is the average Hessian of the negative log posterior. Taking the exponential, the right hand side of eq. \ref{eq:taylor} is of Gaussian functional form, resulting in 
\begin{equation}\label{eq:posterior}
    \mathbf{W}\sim\mathcal{N}(\mathbf{W}^\star,H^{-1}).
\end{equation}
\begin{figure}[t]
    \centering
    \begin{subfigure}{.435\linewidth}
        \centering
        \includegraphics[width=\linewidth]{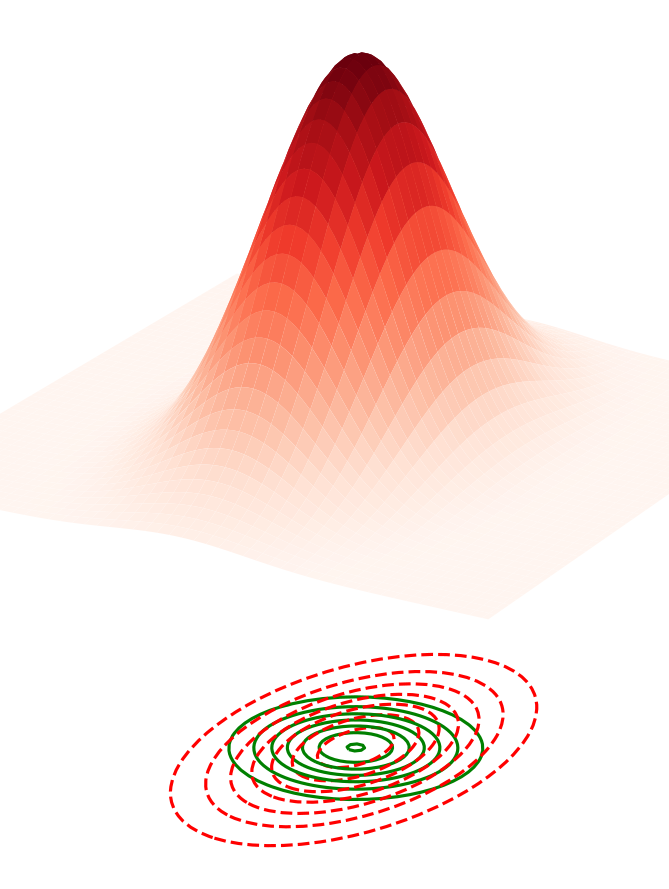}
        \caption{No regularization}
        \label{fig:no_reg}
    \end{subfigure}
    \begin{subfigure}{.545\linewidth}
        \centering
        \includegraphics[width=\linewidth]{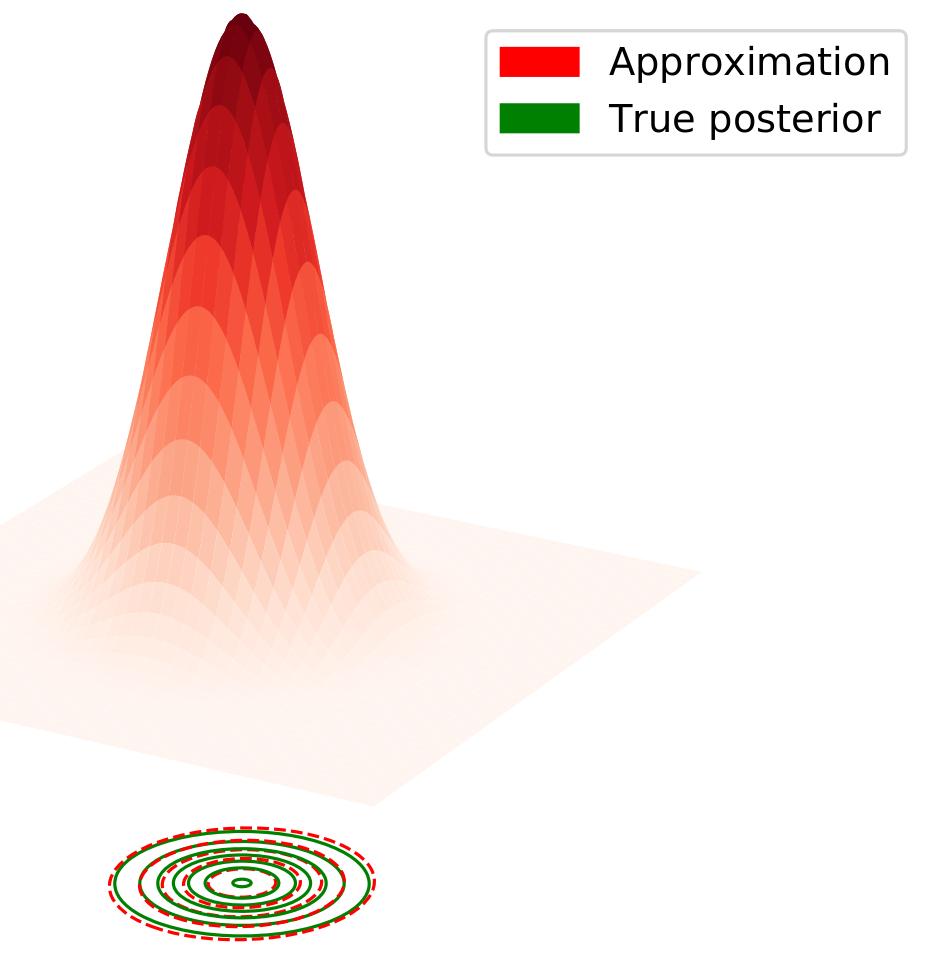}
        \caption{Regularized}
        \label{fig:reg}
    \end{subfigure}
    \caption{An illustration showing how Bayesian Optimization can help to remove the under-fitting problem where LA puts high probability mass on areas of low true probability (left). This is achieved by regularizing the covariance matrix by optimizing the hyperparameters $N$ and $\tau$ so that the under-fitting areas are removed (right).}
    \label{fig:ill:method}
\end{figure}

Sampling from this posterior allows to perform approximate Bayesian inference for an unseen datum $\mathbf{x}^\star$ using Monte Carlo integration
\begin{align}\label{eq:inference}
    p(\mathbf{y}^\star|\mathbf{x}^\star,\mathcal{D})&=\int p(\mathbf{y}^\star|\mathbf{x}^\star,\mathbf{W})p(\mathbf{W}|\mathcal{D})\mathrm{d}\mathbf{W}\\
    &\approx\frac{1}{T}\sum_{t=1}^T p(\mathbf{y}^\star|\mathbf{x}^\star,\mathbf{W}_t),
\end{align}
where $\mathbf{W}_t$ are weight samples from eq. \ref{eq:posterior}. Note that those samples can be drawn efficiently using a matrix normal distribution \cite{ritter2018scalable}.
Because the Hessian of a NN is in general too large to be computed, stored and inverted, it is usually approximated by the diagonal or block-diagonal and Kronecker factored (KFAC) \cite{martens2015optimizing} Fisher information matrix or the Generalized Gauss-Newton matrix \cite{botev2017practical}. For generality we will refer to any of these second-order matrices as \textit{curvature matrices} $C_\ell$ for $\ell\in(0,L)$ layers of the DNN.

Several problems can arise using this posterior approximation. (1) The curvature matrix might not be positive semi-definite (p.s.d), so it cannot be inverted. (2) The approximations that made the problem tractable might lead to an overestimation of the variance of the Gaussian distribution in certain directions in parameter space. (3) The approximation of the NNs posterior using a multivariate Gaussian distribution itself might place probability mass in low probability areas of the true posterior \cite{ritter2018scalable}. See Fig. \ref{fig:ill:method}.

To mitigate those problems, regularization can be used. Following \cite{ritter2018scalable}, we make use of the simple regularization scheme
\begin{equation}\label{eq:regularization}
    C=N\mathbb{E}\left[-\frac{\partial\log p(\mathcal{D}|\mathbf{W})}{\partial\mathbf{W}^2}\right]+\tau I,
\end{equation}
where $\tau$ can be interpreted as the precision of a Gaussian prior on the weights of the DNN, equivalent to $L_2$-regularization, and $N$ the size of the dataset, but they can also simply be treated as hyperparameters to be optimized w.r.t. to a designated performance metric on a validation set.

Note, that a single set of regularization parameters is used for all $C_{\ell\in L}$, so one might wonder if maintaining multiple sets of hyperparameters for, e.g., layers with curvature of dissimilar magnitude, could result in superior performance. We will come back to this in section \ref{sec:3:c}.

\subsection{Bayesian Optimization}
\textit{Bayesian Optimization} (BO) is a form of sequential, model-based optimization scheme for finding the minimum of an expensive to evaluate and potentially noisy black-box function $q(r)$. Black-box functions means, that we do not have access to gradient information and the function is too expensive to evaluate to compute numerical derivatives. Instead, $q(r)$ is approximated by a probabilistic model, trained alongside the optimization procedure. The next best candidate $r_{n+1}$ in the search space $\mathcal{R}$ to be evaluated by $q(r)$ is found by first computing the analytical posterior distribution of the probabilistic model and then minimizing an easy to evaluate \textit{utility function} $u(r)$ based on this posterior, with the goal of finding an optimal point $r^\star$. More formally we get
\begin{align}
    r_{n+1}&=\arg\min_{r\in\mathcal{R}}u(r)\\
    r^\star&=\arg\min_{r\in\mathcal{R}}q(r).
\end{align}

 Typical choices for $u$ include \textit{Expected Improvement} (EI) or \textit{Lower Confidence Bound} (LCB) while a common choice of model is the \textit{Gaussian Process} (GP) \cite{shahriari2015taking}.
This optimization procedure stands in contrast to the commonly used grid or random search, where candidate points are either predefined (grid search) or sampled randomly from some predefined distribution (random search).

An typical example in the BO domain is DNN training, where the next best set of hyperparameters--corresponding to the next best candidate $r_{n+1}$--like the learning rate, regularization strength or dropout rate, cannot be directly inferred using e.g. gradient information and the performance of the chosen hyperparameters can only be evaluated after hours, days or even weeks of training, making it an extremely expensive black-box function $q(r)$.

\subsection{Bayesian Optimization meets Laplace Approximation}
\label{sec:3:c}
We face the same problem when searching for an optimal set of regularization parameters for LA, i.e. $r=(N,\tau)$. As can be seen from eq. \ref{eq:inference}, $T$ forward passes through the network need to be performed for each data point in the dataset where larger $T$ yield more accurate approximations. Using DNNs and a large datasets, evaluating the performance of one set of $N$ and $\tau$ is a significant computational burden and is further complicated due to noisy results stemming from the stochastic sampling procedure of LA. However, as explained earlier, the regularization parameters have a defining influence on the performance of LA, as excessive regularization reduces the probabilistic network output to a deterministic one while too little regularization hurts the networks accuracy.

The problem is further aggravated when separate sets of hyperparameters for different layers are introduced, as we are operating under the curse of dimensionality. We therefore propose, with a particular focus on the aforementioned multiple-set case, the use of BO of LA hyperparameter search. In particular, we make use of a GP model in conjunction with the \textit{Hedge} utility function, which chooses probabilistically between standard choices like EI and LCB \cite{shahriari2015taking}, to find a single or multiple sets of regularization parameters for LA.

This is done by an independent optimization of the three utility functions EI, LCB and \textit{Probability of Improvement} (PI), each returning a candidate for the next evaluation $n+1$. The best proposal is picked using a normalized and uniformly initialized weight vector $\mathbf{w}^n$
\[p(\mathbf{w}^n)=\frac{\mathbf{w}^n}{w_1^n+w_2^n+w_3^n},\]
which is updated using cost $c_i^n$ for utility function $i\in\{1,2,3\}$ at evaluation $n$ from the chosen performance metric as
\[w_i^{n+1}=w_i^n\cdot\beta^{c_i^n},\]
where $\beta\in[0,1]$ is a hyperparameter of the Hedge algorithm \cite{freund1995desicion}.

The most common performance metrics for classification are accuracy and negative log-likelihood (NLL), i.e. the average cross entropy of the target class $y_{ik}=1$ for sample $i$ and the predicted probability $\hat{y}_{ik}$, though they are insufficient to capture important notions when reliable uncertainty estimates are required. As was shown in \cite{guo2017calibration}, modern DNNs are severely miscalibrated, providing accurate yet highly overconfident predictions where the predictive uncertainty does not reflect the empirical frequency of being right or wrong. And while NLL captures notions of uncertainty in the form of predictive entropy, it does not account for the fact that DNNs can overfit w.r.t. classification \textit{and} probabilistic error, s.t. NLL increases while accuracy further improves. Next to accuracy, we therefore make use of \textit{expected calibration error} (ECE) \cite{guo2017calibration} as the (binned) weighted difference between accuracy and confidence, to provide a more faithful representation of the classifiers performance, ignoring additional benefits of BNNs s.a. superior detection capabilities of out-of-domain examples or increased resilience to adversarial attacks \cite{ritter2018scalable} in this discussion for the sake of brevity. We subsequently provide definitions of the previously introduced performance metrics for reference.
\section{EXPERIMENTS}
To empirically validate the effectiveness, efficiency and scalability of BO for LA we conduct a small scale comparison of BO and random search on the CIFAR-10 \cite{krizhevsky2009learning} dataset and a large scale experiment on ImageNet \cite{deng2009imagenet}, using publicly available pre-trained networks to facilitate reproducibility. All experiments where implemented using PyTorch in Python and we make the code publicly available.

\subsection{Small scale experiment}
In the small scale experiment we make use of a ResNet variant \cite{he2016deep} with 18 layers (ResNet18), pre-trained on CIFAR-10\footnote{Publicly available at \url{https://github.com/huyvnphan/PyTorch-CIFAR10}} to $93.32\%$ validation accuracy (and therefore conversely an error of $6.68\%$) on random $50\%$ of the official validation set, which we will use for the hyperparameter optimization. The ECE is $4.57\%$ resulting in a baseline cost of $11.25$ as the sum of classification error and ECE which we will use as performance measure.

We use our own LA implementation with $T=30$ posterior samples from the diagonal posterior approximation computed using the CIFAR-10 training set with the same augmentation techniques as applied during network training\footnote{We opted for the simple diagonal approximation as opposed to more sophisticated approximations like KFAC as our focus lies on the benefits of using BO for LA and not on achieving state-of-the-art performance.}.

\begin{figure}[thpb]
    \centering
    \includegraphics[width=\linewidth]{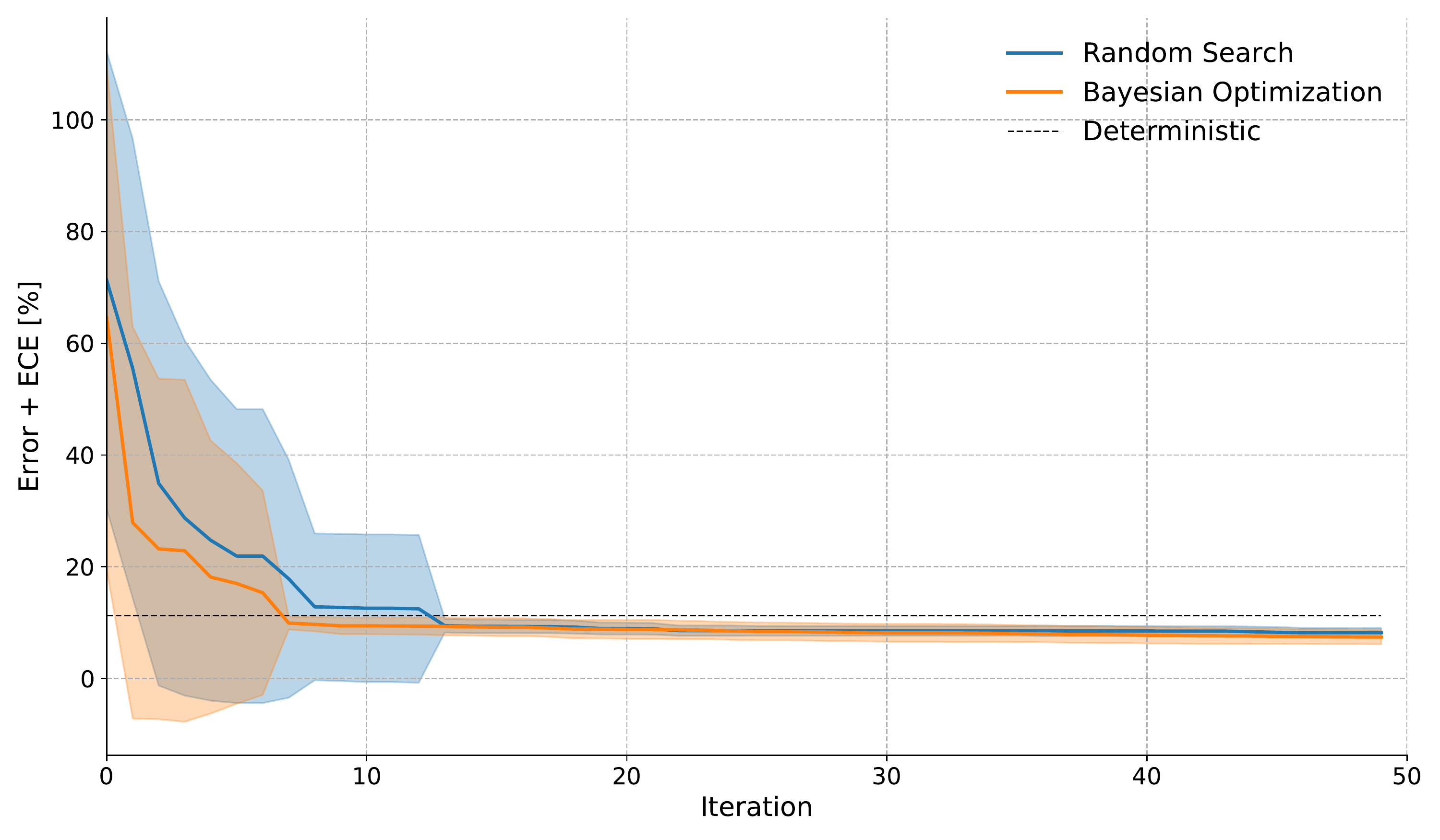}
    \caption{Average performance of BO and RS for 50 evaluations using ResNet18 on CIFAR-10 with 2 hyperparameters. We note that BO converges faster than random search which is evident in the transient phase (i.e. before iteration 15). This shows that the proposed BO can provide a more powerful scheme to optimize calibration and accuracy simultaneously.}
    \label{fig:small1}
\end{figure}

Figure \ref{fig:small1} shows the lowest cost found by BO and RS for 50 evaluations averaged over 20 runs (solid line) and one standard deviation (shaded area) when searching for one set of hyperparameters for LA.

\begin{figure}[thpb]
    \centering
    \begin{subfigure}{.49\linewidth}
        \centering
        \includegraphics[width=\linewidth]{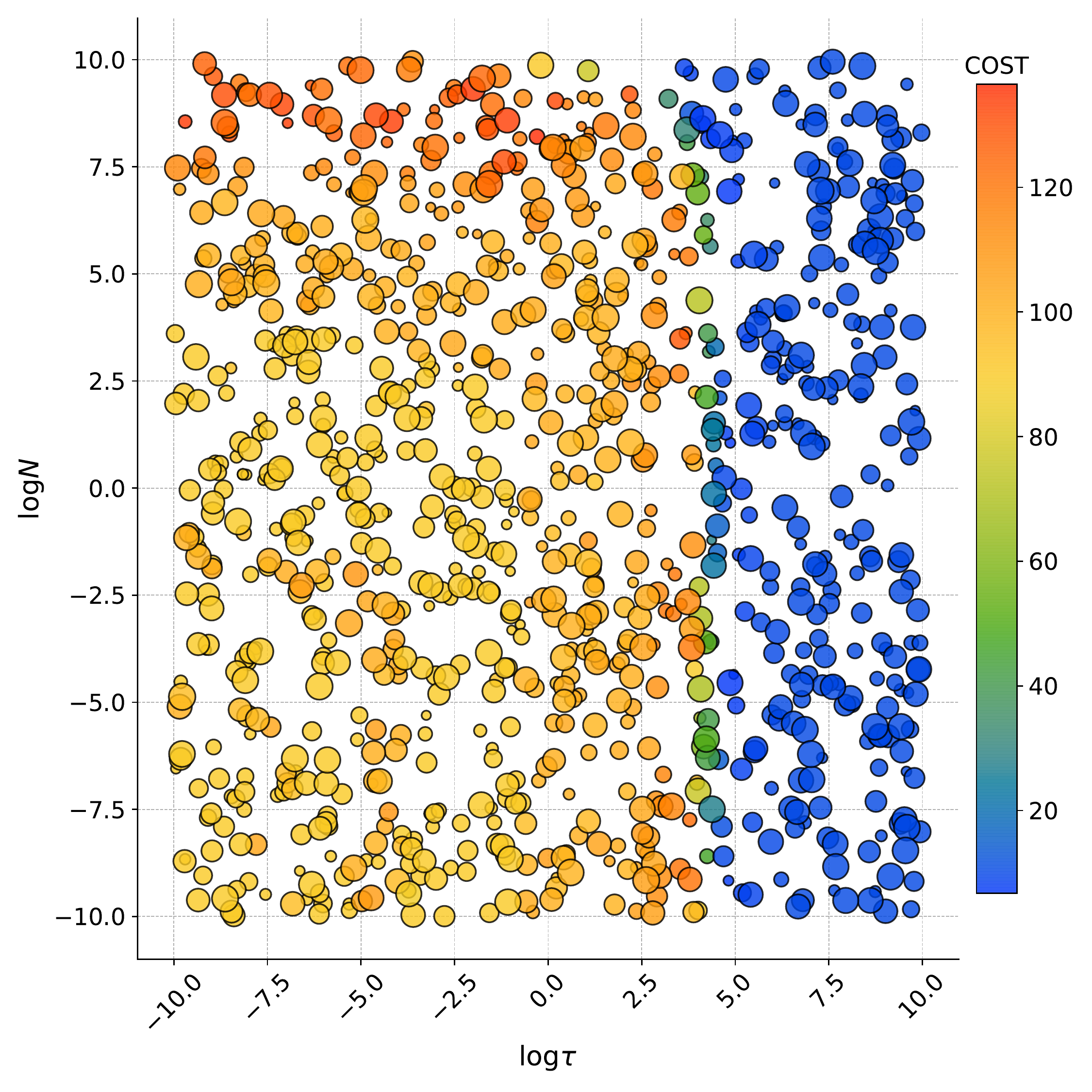}
        \caption{Sampling behavior of RS.}
        \label{fig:small2a}
    \end{subfigure}
    \begin{subfigure}{.49\linewidth}
        \centering
        \includegraphics[width=\linewidth]{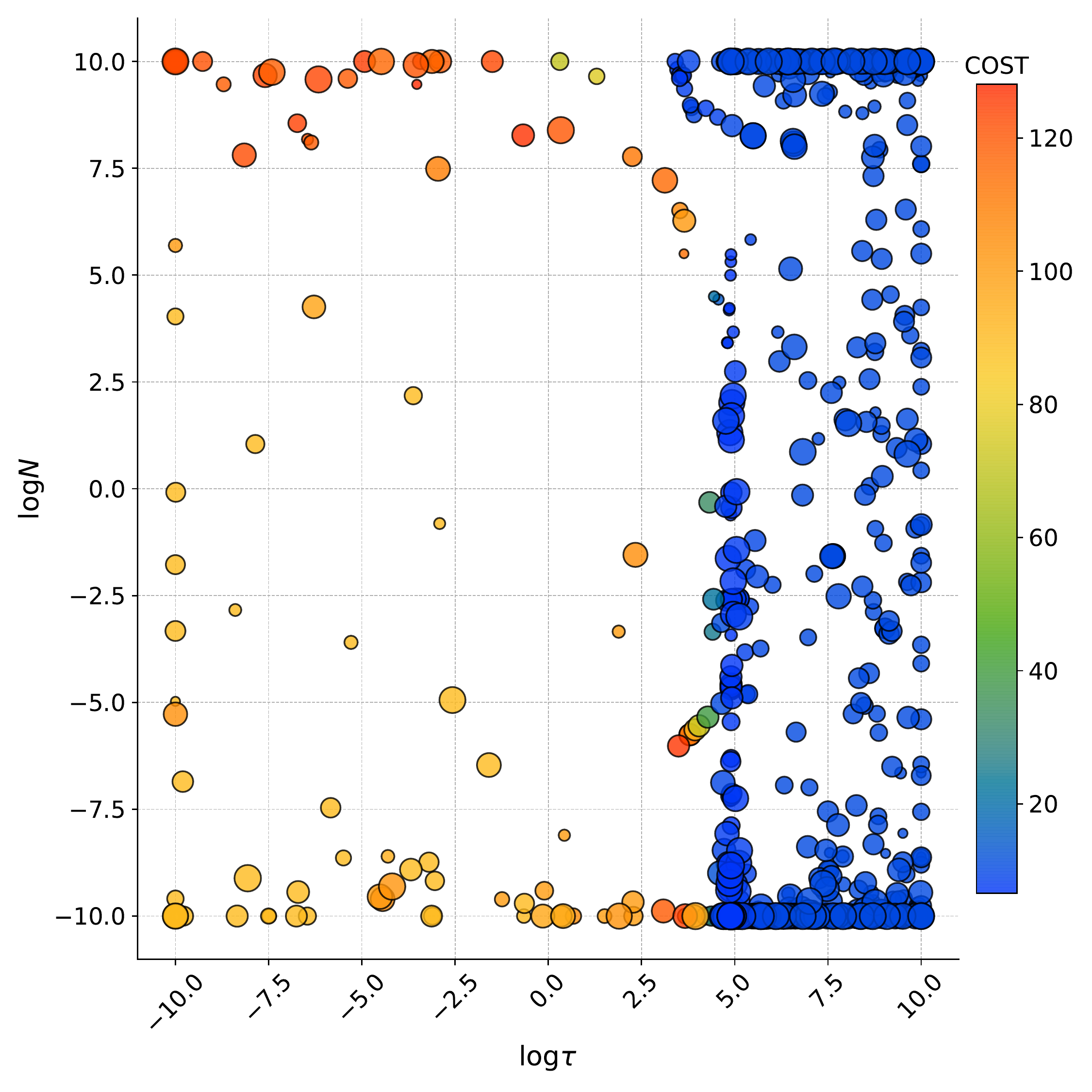}
        \caption{Sampling behavior of BO.}
        \label{fig:small2b}
    \end{subfigure}
    \caption{While RS samples the search space uniformly at random, BO focuses the search on certain areas, largely avoiding the high cost area to the left. High cost is depicted from red to orange while low cost is shown in blue. This demonstrates the efficiency of the proposed algorithm in improving the under-fitting problem of Laplace Approximation.}
    \label{fig:small2}
\end{figure}

As can be observed, BO for LA is able to surpass the performance of the standard DNN (black dashed line) after seven evaluations compared to 13 for RS and additionally finds a marginally superior final set of parameters. We can further see, that the variance of the RS performance is much higher for a small number of evaluations, making it a less reliable choice for a high cost setting. We additionally visualize the behavior of both algorithms by showing their sampling strategies side by side (Fig. \ref{fig:small2}). Clearly, BO makes better use of its sampling budget by largely excluding the high cost area of insufficient regularization.

\begin{figure}[thpb]
    \centering
    \includegraphics[width=\linewidth]{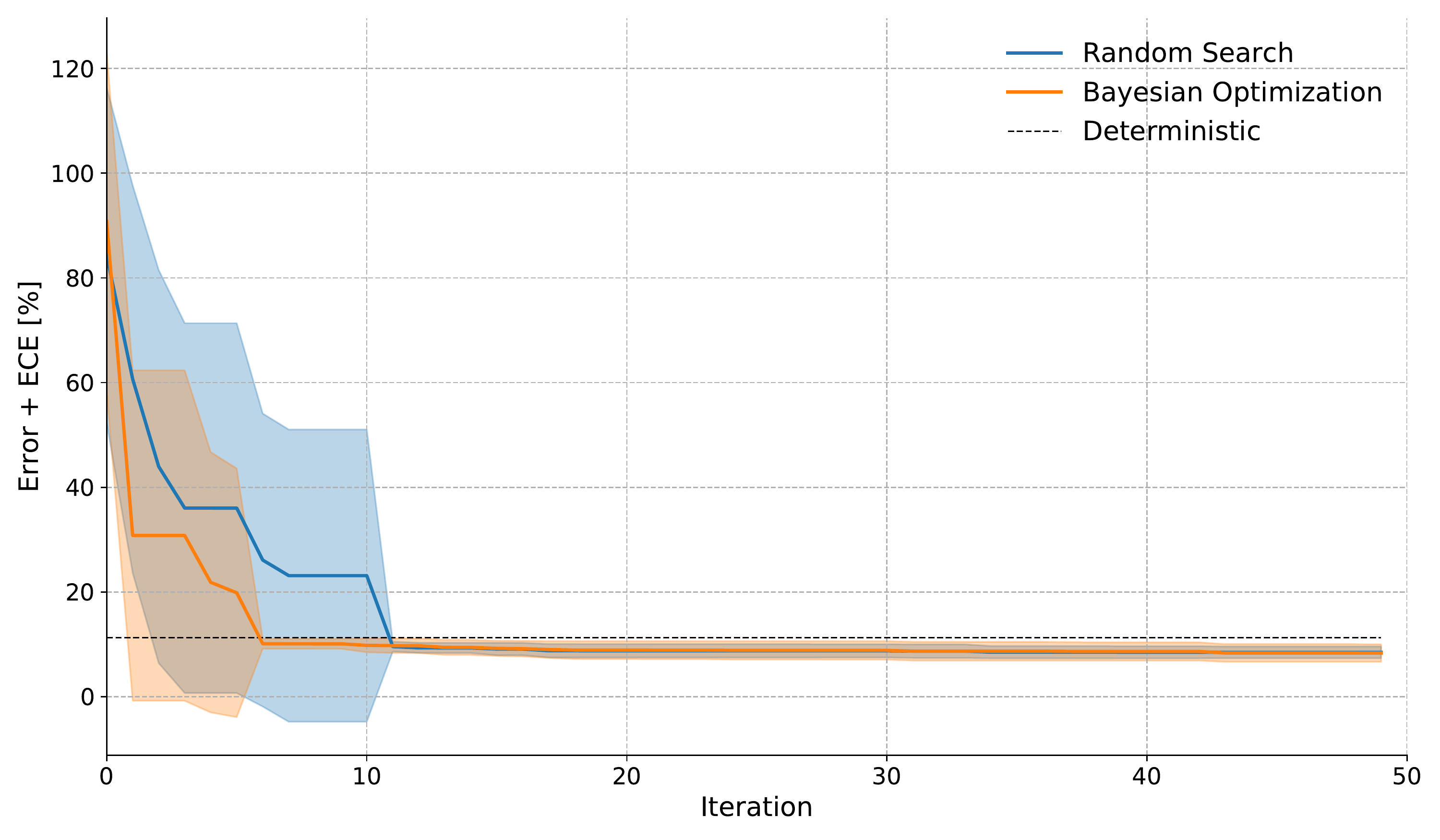}
    \caption{Average performance of BO and RS for 50 evaluations using ResNet18 on CIFAR-10 with 4 hyperparameters. As the number of hyperparameters grow, and as we treat to regularize the posterior of each layers separately, BO further outperforms random search w.r.t convergence rate.}
    \label{fig:small3}
\end{figure}

We conduct an additional small scale experiment with the exact same setup as before except for the introduction of an additional set of hyperparameters. The first set is used to regularize the convolutional layers while the second set regularizes the final fully connected layer. While this is mostly done to highlight the benefit of BO over RS in higher dimensional search spaces, it is reasonable to expect performance gains from regularizing individual layers differently, as the magnitude of their curvature matrices varies significantly.
The results are shown in Fig. \ref{fig:small3}. Again, BO converges faster than RS and is able to find slightly superior sets of hyperparameters. The results are summarized in table \ref{table:small}.

It should be noted that, even though we did not strive for state-of-the-art performance by making use of the rather simplistic diagonal curvature approximation, LA could significantly improve the performance of ResNet18 on CIFAR-10, with a slight improvement in accuracy and a major reduction of mis-calibration, which might not be immediately visible from the preceding figures.

\begin{table}[thpb]
\caption{Small scale results after convergence. Higher is better for accuracy (Acc) while lower is better for expected calibration error (ECE). After convergence, the combination of LA and BO achieves better calibration performance. Furthermore, accuracy of the predictions has improved over the deterministic counterpart due to the marginalization over the possible weights.}
\label{table:small}
\begin{center}
\begin{tabular}{|c|c|c|c|c|c|c|}
\hline
\multicolumn{1}{|c|}{} & \multicolumn{2}{c|}{\textbf{Baseline}} & \multicolumn{2}{c|}{\textbf{LA + RS}} &
\multicolumn{2}{c|}{\textbf{LA + BO}}\\
\cline{2-7}
 & Acc. & ECE & Acc. & ECE & Acc. & ECE \\
\hline
2 parameters & 93.32 & 4.57 & 93.68 & 0.41 & 93.6 & \textbf{0.18}\\
\hline
4 parameters & 93.32 & 4.57 & 93.8 & 0.7 & 93.62 & \textbf{0.17}\\
\hline
\end{tabular}
\end{center}
\end{table}

\subsection{Large scale experiment}
We finally conduct a large scale experiment on ImageNet where BO is especially useful, because every single performance evaluation of the chosen LA hyperparameters is extremely costly. We change the network architecture to DenseNet121 \cite{huang2017densely} which comes pre-trained on ImageNet as part of the \textit{torchvision} PyTorch module and achieves a baseline accuracy of $74.42\%$ and ECE of $2.63\%$ on a randomly chose $50\%$ subset of the official ILSVRC-2012 validation set. The other half is used to perform the LA hyperparameter optimization.
Guided by the results from the previous experiments we define a fixed search budget of 10 evaluations for both RS and BO.

\begin{figure}[thpb]
    \centering
    \begin{subfigure}{.49\linewidth}
        \centering
        \includegraphics[width=\linewidth]{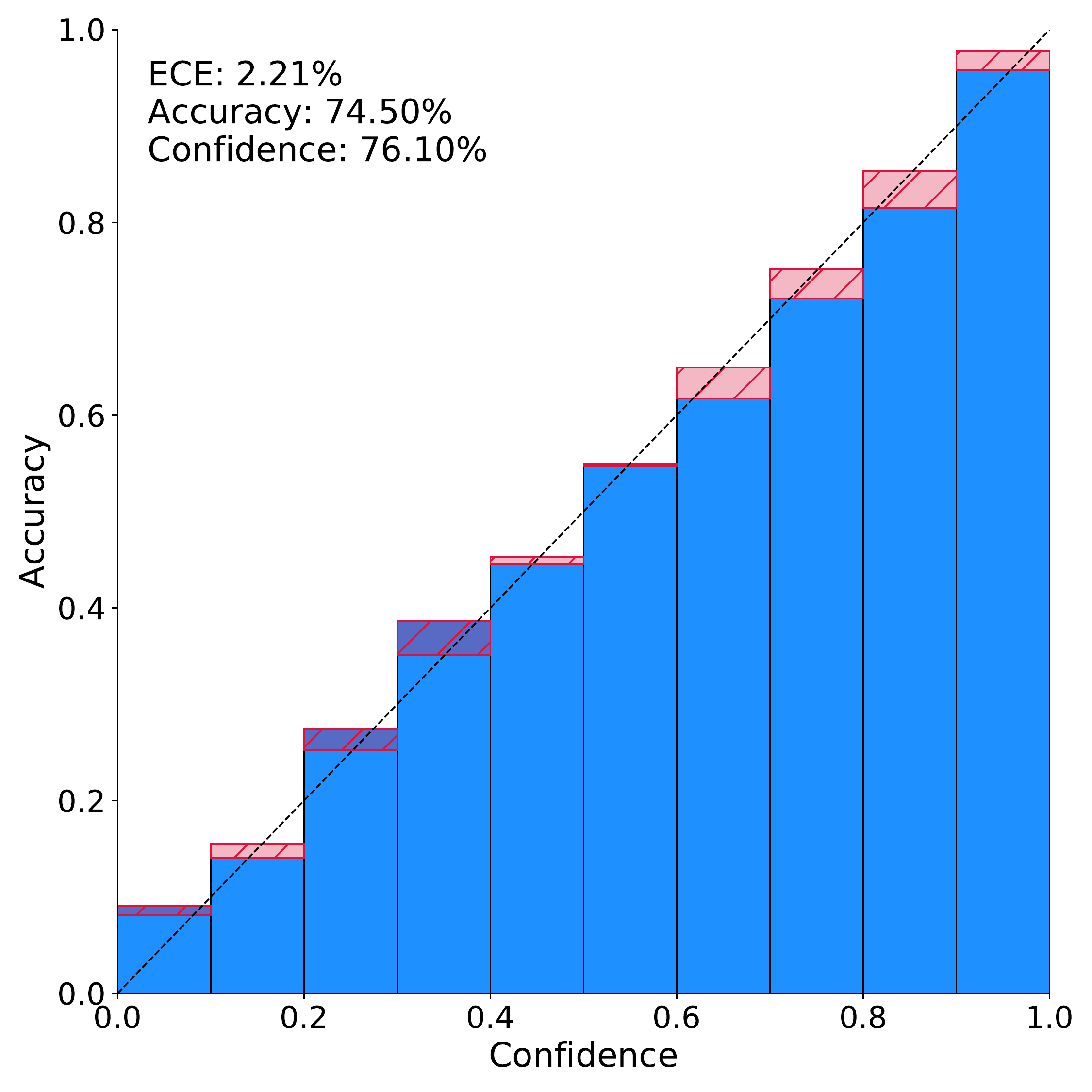}
        \caption{Random Search}
        \label{fig:large_rs}
    \end{subfigure}
    \begin{subfigure}{.49\linewidth}
        \centering
        \includegraphics[width=\linewidth]{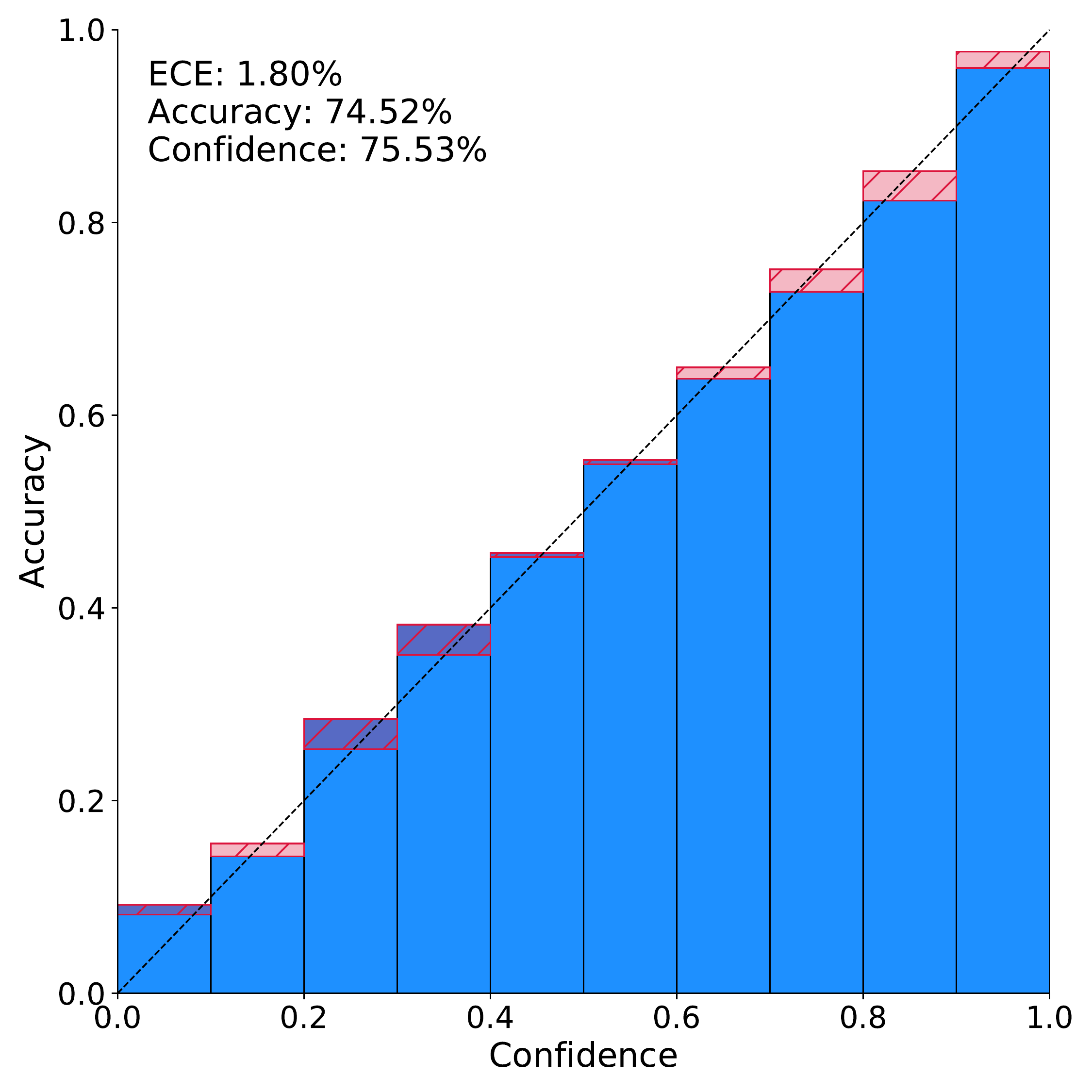}
        \caption{Bayesian Optimization}
        \label{fig:large_gp}
    \end{subfigure}
    \caption{Reliability diagrams for LA of DenseNet121 with hyperparameter optimization using RS and BO.}
    \label{fig:large}
\end{figure}

The results are depicted in figure \ref{fig:large} in the form of two \textit{Reliability Diagrams} inspired by \cite{guo2017calibration}, which visualize accuracy as a function of confidence by showing the average accuracy per confidence bin as blue bars, and their difference (the calibration gap) as red bars on top. A perfectly calibrated network exhibits the same accuracy in each bin as the corresponding confidence dictates, resulting in the identity function, indicated by the dashed black line.

Even though both methods find regularization parameters that outperform the deterministic network, BO achieves lower ECE and slightly higher accuracy (see figure labels) within the given optimization budget. As the deterministic network was already relatively well calibrated (ECE of $2.63\%$), absolute performance gains can mostly be seen for confidence bin $0.6-0.7$, where the BO results show significantly lower overconfidence.
\section{CONCLUSION}
In this work, we have presented a practical approach based on Laplace Approximation for estimating the model uncertainty in deep learning methods, and further proposed to use a novel Bayesian optimization technique to regularize the obtained representation of model uncertainty to better fit the ground truth. We demonstrated both performance and scalability by outperforming random search as a strong baseline and further demonstrated the relevancy in robotics by showing the scalability of our approach on the ImageNet dataset using a large deep neural network.


%%%%%%%%%%%%%%%%%%%%%%%%%%%%%%%%%%%%%%%%%%%%%%%%%%%%%%%%%%%%%%%%%%%%%%%%%%%%%%%%

%%%%%%%%%%%%%%%%%%%%%%%%%%%%%%%%%%%%%%%%%%%%%%%%%%%%%%%%%%%%%%%%%%%%%%%%%%%%%%%%

%%%%%%%%%%%%%%%%%%%%%%%%%%%%%%%%%%%%%%%%%%%%%%%%%%%%%%%%%%%%%%%%%%%%%%%%%%%%%%%%
%\section*{APPENDIX}


%\section*{ACKNOWLEDGMENT}

%The preferred spelling of the word acknowledgment in America is without an e after the g. Avoid the stilted expression, One of us (R. B. G.) thanks . . .  Instead, try R. B. G. thanks. Put sponsor acknowledgments in the unnumbered footnote on the first page.

%%%%%%%%%%%%%%%%%%%%%%%%%%%%%%%%%%%%%%%%%%%%%%%%%%%%%%%%%%%%%%%%%%%%%%%%%%%%%%%%

%References are important to the reader; therefore, each citation must be complete and correct. If at all possible, references should be commonly available publications.

\bibliographystyle{IEEEtran}
\bibliography{references}

\end{document}